\documentclass[10pt,twocolumn,letterpaper]{article}

\usepackage[final]{cvpr}      
\usepackage[table]{xcolor} 
\usepackage{multirow}
\definecolor{cvprblue}{rgb}{0.21,0.49,0.74}
\usepackage[pagebackref,breaklinks,colorlinks,allcolors=cvprblue]{hyperref}

\def\paperID{*****} 
\def\confName{CVPR}
\def\confYear{2026}

\title{Nighttime Hazy Image Enhancement via Progressively and Mutually Reinforcing Night-Haze Priors}
\author{
Chen Zhu\quad
Huiwen Zhang\quad
Mu He\quad 
Yujie Li\quad 
Xiaotian Qiao\footnotemark[2]
\\
\\
 Xidian University
}

\begin{document}
\maketitle

\renewcommand{\thefootnote}{\fnsymbol{footnote}}
\footnotetext[2]{Corresponding author.}

\begin{abstract}
Enhancing the visibility of nighttime hazy images is challenging due to the complex degradation distributions. Existing methods mainly address a single type of degradation (e.g., haze or low-light) at a time, ignoring the interplay of different degradation types and resulting in limited visibility improvement. We observe that the domain knowledge shared between low-light and haze priors can be reinforced mutually for better visibility. Based on this key insight, in this paper, we propose a novel framework that enhances visibility in nighttime hazy images by reinforcing the intrinsic consistency between haze and low-light priors mutually and progressively. In particular, our model utilizes image-, patch-, and pixel-level experts that operate across visual and frequency domains to recover global scene structure, regional patterns, and fine-grained details progressively. A frequency-aware router is further introduced to adaptively guide the contribution of each expert, ensuring robust image restoration. Extensive experiments demonstrate the superior performance of our model on nighttime dehazing benchmarks both quantitatively and qualitatively. Moreover, we showcase the generalizability of our model in daytime dehazing and low-light enhancement tasks.
\end{abstract}

\section{Introduction}
\label{sec:intro}
\begin{figure}[t]
\centering
\includegraphics[width=\linewidth]{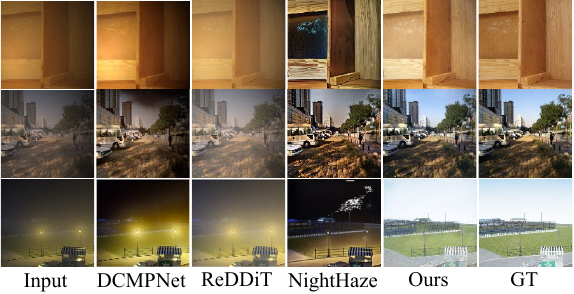}
\caption{
Comparisons with recent state-of-the-art methods (DCMPNet~\cite{DCMPNet} for daytime dehazing, ReDDiT~\cite{ReDDiT} for low light enhancement, NightHaze~\cite{NightHaze} for nighttime dehazing) and our method on diverse nighttime hazy scenarios.
Conventional methods that address either daytime dehazing or low-light enhancement in isolation can only recover limited aspects of nighttime hazy images. In contrast, our method leverages both priors jointly, achieving more comprehensive image restoration and surpassing the performance of existing state-of-the-art nighttime dehazing techniques.
}
\vspace{-0.1in}
\label{intro}
\end{figure}

Haze is a common adverse environmental condition that degrades human perception of visual content and adversely affects downstream multimedia applications like security and surveillance.
Such a phenomenon can be particularly problematic at night, where the interplay of low light and dense haze severely obscures scene details.
While significant progress has been made in daytime image dehazing and low-light image enhancement~\cite{dehaze4,dehaze5, liu2024hcanet, Retinexformer, ReDDiT, lle1, RetinexNet, lle5}, as shown in the 2nd and 3rd columns of Figure~\ref{intro}, these approaches are tailored to specific degradation types and fail to address the unique low-visibility challenges of nighttime hazy scenarios.

To tackle the diverse degradations in nighttime haze images, model-based methods~\cite{MPRNet, Gray_Haze-Line_Prior, TANG2021103086} have been proposed, relying on various prior hypotheses and statistical laws for haze removal.
However, these approaches focus narrowly on specific corruption factors, failing to address the full spectrum of degradation types in nighttime hazy scenarios.
Recent deep learning-based approaches\cite{HDP-Net, NightHaze, NightEnhance, SFSNID} offer improvements but remain constrained by the absence of real-world paired training data.
As illustrated in the 4th column of Figure~\ref{intro}, training on synthetic or unpaired real data yields suboptimal visibility enhancement results, due to the considerable domain gap between synthetic and real-world degradations.
Moreover, these methods lack the generalizability to learn balanced priors for handling multiple degradations simultaneously in nighttime hazy scenarios.

From our study, we have investigated the distinctive characteristics of nighttime haze images and made the following key insights.
Nighttime haze images represent complex combinations of low light and dense haze conditions.
From the visual domain perspective, low-light skews pixel intensities toward black with low-intensity values, while haze shifts toward white with high-intensity pixels, yet both strive for the uniform intensity distribution of clear images.
From the frequency domain perspective, dehazing and low-light enhancement share a common goal of restoring high-frequency details lost in degraded images.
We hypothesize that explicitly modeling the mutual knowledge between low-light enhancement and dehazing across domains can synergistically enhance visibility in nighttime haze images, leveraging existing paired training data for specific degradation types to further boost performance.

Motivated by the above observations, in this paper, we propose Multi-level Mixture of Mutual Knowledge Experts (M\textsuperscript{3}KE), a novel framework for nighttime haze image enhancement by mutually reinforcing the intrinsic consistency between hazing and low-light priors.
Unlike prior Mixture-of-Experts (MoE) for low-level vision tasks~\cite{WM-MoE, Image_Restoration, Deweahter} that rely on numerous experts without considering the intrinsic relationships, M\textsuperscript{3}KE effectively facilitates the complementary information exchange in nighttime haze images using only a limited number of experts.
In particular, M\textsuperscript{3}KE utilizes image-, patch-, and pixel-level experts that operate across visual and frequency domains to recover global scene structure, regional patterns, and fine-grained details progressively. 
Each expert incorporates an Integrated Frequency Interaction Block (IFIB), which fuses visual and frequency information to prioritize degradation-relevant channels while suppressing irrelevant ones.
Additionally, a frequency-aware router (FAR) is introduced to adaptively assign weights to these experts across multiple levels based on input characteristics, ensuring robust adaptation to diverse conditions.

The extensive experimental results show that M\textsuperscript{3}KE achieves state-of-the-art performances on the nighttime dehazing task quantitatively and qualitatively.
We further conduct experiments on the daytime dehazing and the low-light enhancement tasks, underscoring the generalizability of M\textsuperscript{3}KE in addressing visibility enhancement for diverse hazy and low-light scenarios. 
The main contributions of this paper are summarized as follows:
\begin{itemize}
\item We make the first attempt to explicitly consider the intrinsic consistency between hazing and low-light priors for visibility enhancement in nighttime haze scenarios.
\item We propose M\textsuperscript{3}KE, a Multi-level Mixture of Mutual Knowledge Experts framework to learn complementary information exchange in nighttime haze images from visual and frequency domains with efficient experts.
\item The experimental results show M\textsuperscript{3}KE’s superior visibility enhancement for nighttime haze images and generalizability across diverse hazy and low-light scenarios.
\end{itemize}

\begin{figure*}[t]
\centering
\includegraphics[width=1.99\columnwidth]{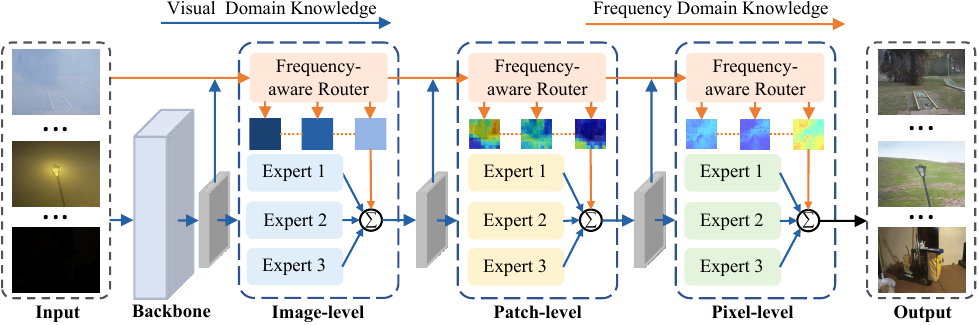} 
\caption{
The overall pipeline of our M\textsuperscript{3}KE framework.
Given a degraded image under diverse hazy or low-light scenarios as input, M\textsuperscript{3}KE progressively enhance the visibility through image-, patch-, and pixel-level experts respectively.
A Frequency-aware Router is employed to guide the weight assignment for each expert.
Finally, a clear image is generated as the output.}
\label{Overallpipline}
\end{figure*}

\begin{figure}[t]
\centering
\includegraphics[width=0.99\columnwidth]{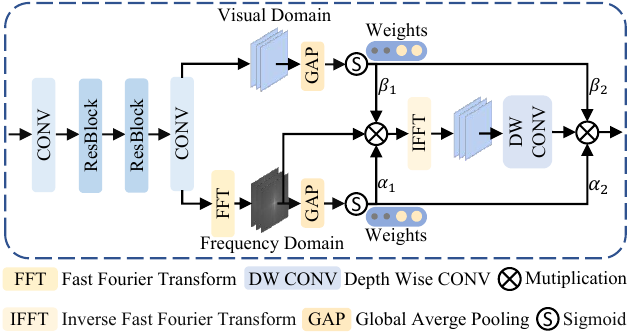} 
\caption{Illustration of the Expert Block, which is designed to fuse and enhance features from different domains.}
\label{IFIB}
\end{figure}

\section{Related Work}
\paragraph{Daytime Image Dehazing.}
Daytime haze scenarios are characterized by uniform atmospheric light, typically dominated by skylight, enabling a range of effective dehazing algorithms.
Early methods typically rely on atmospheric scattering models, which assume that the lighting is uniform and originates from the sky, to perform dehazing on daytime haze images. 
A variety of priors, such as the dark channel prior (DCP) \cite{DCP} and the color attenuation prior (CAP) \cite{CAP}, are used to estimate the transmission map and atmospheric light, recovering the clear image by inversely solving the atmospheric scattering model.
However, the performance of these methods heavily depends on the accurate estimation of the transmission map and atmospheric light, resulting in poor performance in real-world scenarios.
End-to-end neural networks and attention-based techniques have shown significant success in dehazing tasks~\cite{AOD, MSCNN, dehaze1, dehaze2, dehaze3, islam2024hazespace2m, liu2021synthetic, liu2024hcanet, MITNet}. 
Recently, DEA-Net~\cite{DEANet} enhances the network’s feature representation capability through structural optimization and attention-based feature fusion, achieving a favorable balance between accuracy and efficiency.
DCMPNet~\cite{DCMPNet} exploits depth cues to excel on synthetic benchmarks.

Although the aforementioned methods show significant progress for daytime image dehazing, they fail to achieve quality improvement for nighttime hazy scenarios that contain multiple complex degradation types.
In contrast, our M\textsuperscript{3}KE framework leverages the intrinsic consistency between hazing and low-light priors, superior performance across both daytime and nighttime dehazing tasks.

\paragraph{Nighttime Image Dehazing.}
Nighttime dehazing \cite{MPRNet, NightVDM, NightDeFog, NightDehaze,NightHaze} is inherently more challenging than daytime dehazing due to the interplay of dense haze with low illumination, uneven lighting, halo effects, and color distortions.
Early model-based methods extended atmospheric scattering models to nighttime scenarios.
For instance, Zhang et al. \cite{Fast_Haze} introduce the maximum reflectance prior that estimates ambient illumination and transmission by utilizing high-intensity patches in each color channel. However, these approaches struggled with the multifaceted nature of nighttime haze, often addressing only localized issues like halos or low contrast, with limited effectiveness in complex scenes.
Deep learning has driven significant progress in nighttime dehazing. 
Jin et al. \cite{NightEnhance} introduce a halo suppression method combining APSF-guided halo rendering and gradient adaptive convolution to effectively address halo and low-light issues in nighttime images. 
Another notable approach is SFSNiD~\cite{SFSNID}, which employs a semi-supervised framework that integrates spatial-frequency domain interaction and pseudo-label retraining, mitigating glow and unrealistic brightness caused by domain shifts.

While existing supervised or semi-supervised methods have shown promising results for nighttime image haze removal, they struggle to generalize well to real-world scenarios.
A considerable domain gap exists between synthetic and real-world degradations hampers visibility enhancement, while imbalanced priors further prevent effective handling of multiple degradations simultaneously.
To mitigate these issues, our M\textsuperscript{3}KE framework explicitly leverages the mutual knowledge between low-light and hazing priors to synergistically enhance visibility in nighttime haze images, achieving superior performance for diverse hazy and low-light scenarios.

\section{Method}
Our goal is to learn a function $f$ that transforms a degraded nighttime hazy image into a clear, artifact-free output.
The key insight lies in harnessing a multi-level mixture of knowledge experts to mutually reinforce the intrinsic synergy between dehazing and low-light priors, addressing the complex degradations in nighttime haze scenarios.
As shown in Figure~\ref{Overallpipline}, M\textsuperscript{3}KE framework employs a progressive restoration strategy, comprising image-level, patch-level, and pixel-level experts that collaboratively recover global scene structure, regional patterns, and fine details.
A Frequency-Aware Router (FAR) adaptively assigns weights to these experts across multiple levels.
The output at each level is computed as a weighted combination of expert contributions, facilitating a cohesive restoration process.

\subsection{Expert Block}
Each expert in M\textsuperscript{3}KE is designed to fuse and enhance features from the visual and frequency domains.
As illustrated in Figure~\ref{IFIB}, Expert Block starts with a two-layer stacked residual block for initial feature extraction, where each residual layer follows a conv-relu-conv structure, connected through residual connections. 
Subsequently, two parallel branches are used to extract feature vectors from the visual and frequency domains, applying channel attention to enhance features in both domains.
In the visual branch, we apply global pooling and linear mapping with a sigmoid activation directly to the convolutional feature map to obtain the visual-domain feature vector.
In the frequency branch, the input undergoes a Fast Fourier Transform (FFT), with the magnitude of the resulting complex values processed via GAP and a linear layer with sigmoid activation to yield a frequency-domain feature vector.

Both visual and frequency domain feature vectors are then used to apply channel attention to the frequency features obtained via FFT, with the attention balanced by learnable parameters $\alpha$ and $\beta$. 
The enhanced features are subsequently transformed back to the visual domain using an Inverse FFT (IFFT), followed by a depthwise convolution (DWConv) to refine inter-channel relationships.
Finally, channel attention is reapplied to the visual features using the combined domain feature vectors, again balanced by a separate set of learnable parameters $\alpha$ and $\beta$ as follows:
\begin{equation}
w=\operatorname{Sigmoid}(\operatorname{linear}(\operatorname{GAP}(f))),
\end{equation}
\begin{equation}
IFIM_{ out}=\alpha \sum_{i=1}^{C} w_{s}^{i} \cdot c_{i}+\beta \sum_{i=1}^{C} w_{f}^{i} \cdot c_{i},
\end{equation}
where GAP denotes global average pooling, $c_{i}$ is the i-th channel, and $C$ is the total number of channels. $w_{s}^{i}$ and $w_{f}^{i}$ are the weights obtained in the visual and frequency domains. $\alpha$ and $\beta$ are network-learnable parameters used to learn the weight distribution between the visual and frequency domains.

\subsection{Frequency-Aware Router}
To ensure each image is perfectly assigned to its most suitable expert, we propose a Frequency-Aware Router (FAR) that combines both frequency-domain and visual-domain information to determine the assignment weights for each expert. 
Specifically, we first extract features from the intermediate feature map generated by the network, which serves as visual information. 
However, relying solely on visual information may not be sufficient, as the distinction between degraded and restored images diminishes as processing progresses.
Therefore, we apply the Discrete Wavelet Transform (DWT) to the image, capturing low- and high-frequency information.
The frequency-domain knowledge is encoded and passed forward through the network, where it is fused with the visual information from the current feature map to assist FAR in generating accurate expert weights.

\subsection{Multi-level Mixture of Mutual Knowledge}
\paragraph{Image-level.}
The image-level expert initiates coarse restoration by processing initial feature maps extracted via a convolutional backbone.
Specifically, for image-based expert assignment with $n$ experts, the feature map size is $b*c*w*h$, where $b$ is the batch size, $c$ is the number of channels, and $w$ and $h$ represent the width and height, respectively. 
Unlike prior MoE methods that directly apply global pooling, we use two convolutional layers to progressively reduce the size and channel count of the feature map, optimizing computational efficiency. 
We then apply global pooling to obtain a feature vector of size $b*(c//4)*1*1$, followed by a linear layer and softmax to generate expert weights of size $b*n*1*1$. 
The output is a weighted sum of expert contributions as follows: 
\begin{equation}
Output=\sum_{i=1}^{n} G(\operatorname{FAR}(f, d))_{i} \cdot E_{i}(x),
\end{equation}
\begin{equation}
G(x)_{i}=\operatorname{Softmax}(g(x))_{i}=\frac{\exp \left(g(x)_{i}\right)}{\sum_{j=1}^{N} \exp \left(g(x)_{j}\right)},
\end{equation}
where $f$ represents the current feature map, $d$ represents the frequency domain information obtained from DWT, $E_{i}$ represents the $i-th$ expert, and $n$ is the number of experts.

\paragraph{Patch-level.}
The patch-level expert focuses on targeted restoration of image patches, promoting mutual learning across degradation types (e.g., haze, low light).
The frequency-domain feature map, derived via DWT, is similarly partitioned into patches and processed by the FAR to assign expert weights.
After patch restoration, we apply a multi-scale dilated convolution group to expand the receptive field at multiple levels and recover edge features lost during patch division. 
It is followed by a conv-relu residual block, integrated with channel and pixel attention mechanisms to enhance adaptability to local variations.
All operations are integrated using residual connections. 
Note that patches within the same image are assigned to different experts based on local variations, such as dense vs. light haze or strong vs. weak illumination, which is especially beneficial for images with non-uniform degradation. 

\paragraph{Pixel-level.}
The pixel-level expert performs fine-grained refinement, critical for recovering high-frequency details like edges and textures essential to both dehazing and low-light priors.
At this stage, as FAR further extracts features from the fused frequency and visual-domain information to generate expert weights, 
the features are linearly mapped using a $1*1$ convolution to a size of $b*n*w*h$, followed by a softmax to obtain a weight map for each pixel across the $n$ experts.
The final output is a weighted sum of expert contributions, enabling precise restoration of intricate details.
Such a design is crucial for restoring high-frequency information required in both hazing and low-light priors, ensuring effective detail recovery.

\subsection{Training}
During training, we randomly select images from various daytime image dehazing, nighttime image dehazing, and low-light enhancement datasets to form a batch, ensuring robustness across tasks. 
We use a joint loss function $L_{\text{joint}}$, which mainly consists of the Smooth L1 loss $L_{\text{l1}}$, multi-scale structural similarity (MS-SSIM) loss $L_{\text {MS-SSIM}}$, perceptual loss $L_{\text{per}}$, and adversarial loss $L_{\text{adv}}$:
\begin{eqnarray}
L_{\text {joint }}=\lambda_{1} L_{\text{sl1}}+\lambda_{2} L_{\text {MS-SSIM}}+\lambda_{3} L_{\text {per}}+\lambda_{4} L_{\text {adv}},
\end{eqnarray}
where $\lambda_{1}$, $\lambda_{2}$, $\lambda_{3}$ and $\lambda_{4}$ are set to 5, 1, 0.5 and 0.0005, respectively.
Please refer to the supplementary material for more model details.

\begin{figure*}[t]
\centering
\includegraphics[width=2.1\columnwidth]{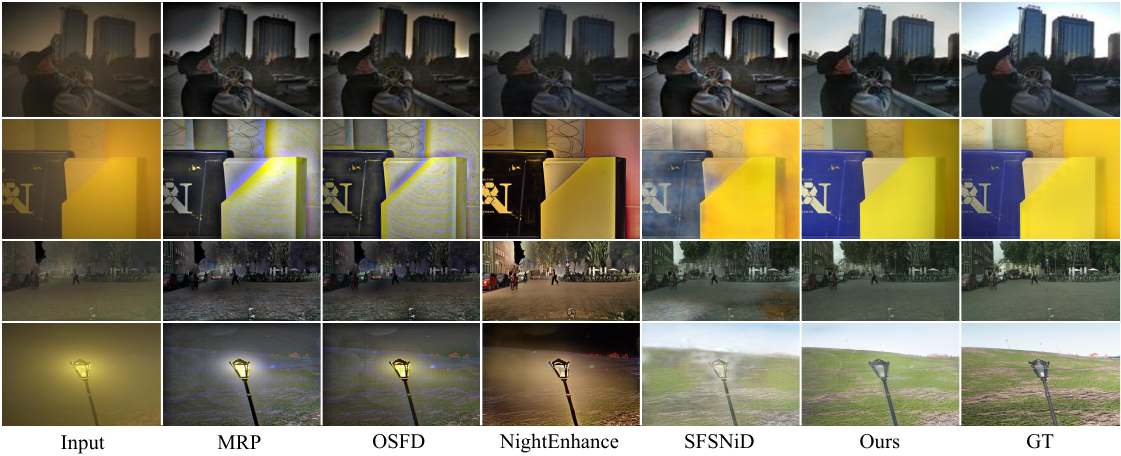} 
\caption{
Qualitative comparison of the proposed method with prior works for synthetic nighttime image dehazing.}
\label{figure_night_time_dehazing}
\end{figure*}

\begin{figure}[t]
\centering
\includegraphics[width=1\columnwidth]{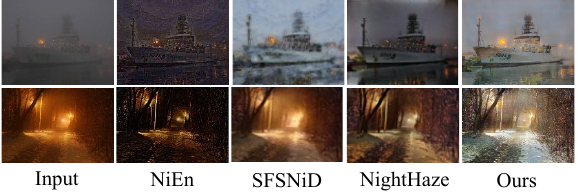}
\caption{
Qualitative comparison of the proposed method with prior works for real-world nighttime image dehazing.}
\label{figure_nighttime_on_real}
\end{figure}

\section{Experiments}
\subsection{Experimental Setup}
\paragraph{Implementation Details.}
All experiments are conducted on an NVIDIA GeForce RTX 4090 using the PyTorch framework.
We optimize M\textsuperscript{3}KE with the Adam optimizer and a cosine annealing learning rate scheduling strategy.
The initial learning rate is set to 5e-5, and the minimum learning rate is set to 5e-7. 
During training, we randomly sample from eighteen datasets covering three priors. For nighttime dehazing, we use NHR \cite{NightDehaze}, NHM \cite{NightDehaze}, NHCL \cite{NightDehaze}, NHCM \cite{NightDehaze}, NHCD \cite{NightDehaze}, and UNREAL-NH \cite{UNREAL}. For the hazy prior, we include Dense-Haze \cite{DENSE-HAZE}, NH-HAZE \cite{NH-HAZE}, O-HAZE \cite{O-HAZE}, I-HAZE \cite{I-HAZE}, SOTS indoor \cite{SOTS}, and SOTS  outdoor \cite{SOTS}. For the low-light prior, we adopt LOL v1 \cite{RetinexNet}, LOLv2-s \cite{LOLV2}, LOLv2-r \cite{LOLV2}, SID-total dark \cite{SID}, and SICE \cite{SICE}, as well as SICE variants like SICE-grad \cite{sice-grad_mixed} and SICE-mixed \cite{sice-grad_mixed}.
Images are randomly cropped to multiples of 256, and then resized to $256*256$ with a batch size of 4 for training, with a total of 400 epochs. 

\paragraph{Datasets.}
We evaluate M\textsuperscript{3}KE on a diverse set of synthetic nighttime dehazing datasets, including NHR \cite{NightDehaze}, NHM \cite{NightDehaze}, NHCL \cite{NightDehaze}, NHCM \cite{NightDehaze}, NHCD \cite{NightDehaze}, and UNREAL-NH \cite{UNREAL}. We also conduct evaluations on real-world nighttime datasets RWNH \cite{NightEnhance}, real-world daytime dehazing datasets Fattal \cite{fattal}, and real-world low-light enhancement datasets, including DICM \cite{DICM}, LIME \cite{LIME}, MEF \cite{MEF}, NPE \cite{NPE}, and VV \cite{VV}.

\begin{table*}
\centering
\caption{Quantitative comparison of nighttime dehazing. The best and second-best results are marked in bold and underlined.}
\label{table_nighttime_dehaze}
\fontsize{8pt}{10pt}\selectfont  
\resizebox{\linewidth}{!}{      
\begin{tabular}{c|cc|cc|cc|cc|cc|cc}
\hline
\multirow{2}{*}{Method} 
& \multicolumn{2}{c|}{NHR} 
& \multicolumn{2}{c|}{NHM} 
& \multicolumn{2}{c|}{NHCL} 
& \multicolumn{2}{c|}{NHCM} 
& \multicolumn{2}{c|}{NHCD}
& \multicolumn{2}{c}{UNREAL-NH} \\ 
\cline{2-13}
& PSNR$\uparrow$ & SSIM$\uparrow$ 
& PSNR$\uparrow$ & SSIM$\uparrow$
& PSNR$\uparrow$ & SSIM$\uparrow$
& PSNR$\uparrow$ & SSIM$\uparrow$
& PSNR$\uparrow$ & SSIM$\uparrow$
& PSNR$\uparrow$ & SSIM$\uparrow$ \\
\hline
GridDehazeNet \cite{Griddehazenet}        & 17.940 & 0.859 & 17.720 & 0.866 & 21.957 & 0.855 & 20.885 & 0.817 & 22.362 &0.877 & 15.698 & 0.633\\ 
FFA \cite{FFA_Net}     & 21.616 & 0.912 & 19.761 & 0.896 & 24.245 & 0.892 & 23.591 & 0.863 & 24.666 & 0.912 & 17.221 & 0.662\\ 

MSBDN \cite{MSBDN}     & 23.504 & 0.939 & 19.117 & 0.890 & 21.505 & 0.881 & 20.528 & 0.848 & 22.171 & 0.904 & 18.677 & 0.683\\ 

DeHamer \cite{Dehamer}   & 23.096 & 0.927 & 20.086 & 0.902 & 24.371 & 0.893 & 23.766 & 0.868 & 24.171 & 0.912 & 16.868 & 0.664\\

C2P \cite{c2p}      & 20.874 & 0.898 & 18.711 & 0.892 & 24.316 & 0.891 & 23.791 & 0.865 & 24.225 & 0.908 & 16.310 & 0.656\\ 

Fourmer \cite{Fourmer}   & 21.952 & 0.905 & 19.485 & 0.887 & 23.886 & 0.875 & 23.381 & 0.851 & 24.233 & 0.893 & 16.113 & 0.645\\

DehazeFormer \cite{DehazeFormer}   & 23.583 & 0.943 & 21.224 & 0.918 & 24.667 & 0.860 & 23.926 & 0.885 & 24.268 & 0.928 & 17.880 & 0.689\\ 

MITNet \cite{MITNet}    & 23.786 &0.941  & 21.734 & 0.914 &23.148  & 0.875 & 23.945 &0.894  & 24.654 & 0.924 & 18.381 & 0.694\\ 

DEANet \cite{DEANet}    & 24.157 & 0.941 & 20.567 & 0.925 & 23.198 & 0.923 & 24.006 &0.899  & 23.999 & 0.894 & \underline{18.791} & 0.701\\ 

CORUN \cite{CORUN}    & 24.997 &0.954  & 20.152 & 0.924 &22.944  & 0.908 & 23.764 &0.913  & 23.487 &0.924  & 18.369 & 0.711\\ 

DCMPNet\cite{DCMPNet}    & 23.756 &0.939  & 22.146 & 0.925 &21.976  & 0.881 & 23.778 &0.909  & 23.475 &0.914  & 17.678 & 0.700\\ 

SGDN\cite{sgdn}    & 24.756 &0.941  & 22.106 & 0.922 &21.819  & 0.876 & 23.318 &0.899  & 23.423 &0.907  & 17.621 & 0.692\\ 

\hline
MRP \cite{Fast_Haze}       &19.931  & 0.777 & 17.741 & 0.710 & 20.022 & 0.685 &20.613 & 0.623 & 17.620 & 0.548 & 15.165 & 0.671\\

OSFD \cite{NightDehaze}      & 21.321 & 0.803 & 19.750 & 0.764 & 20.101 & 0.737 &21.143  & 0.678 & 18.410 & 0.600 & 16.379 & 0.670\\ 

NightDeFog \cite{NightDeFog}& 22.375 & 0.875 &20.561 & 0.891 & 22.156 & 0.907 & 22.157 & 0.857 & 22.189 & 0.901 & 16.146 & 0.699 \\

NightHazeFormer \cite{UNREAL} &24.121 &0.931 & 21.557 &0.910& 23.851 & 0.919 & 23.015 & 0.871 & 23.123&0.921  &17.673  & 0.701 \\

NightEnhance \cite{NightEnhance}&\underline{24.561} &0.941 & 22.157 &\underline{0.924}& 24.357 & 0.921 & 23.975 & 0.911 & 23.999&0.931  &17.973  & 0.721 \\

SFSNiD \cite{SFSNID}& 24.157 & \underline{0.954} & \underline{22.741} & 0.918 & \textbf{24.888} & \underline{0.926} & 23.997 & \underline{0.919} & \underline{24.701} & 0.937 & 17.657 & \underline{0.729} \\

NightHaze \cite{NightHaze}& 24.012 & 0.950 & 21.457 & 0.915 & 23.185 & 0.924 & \underline{24.115} & 0.913 & 24.025 & \underline{0.940} & 18.011 & 0.725 \\

\hline   
\rowcolor{gray!15}  
M\textsuperscript{3}KE (Ours)      & \textbf{25.197} & \textbf{0.962} & \textbf{22.887} & \textbf{0.934} & \underline{24.626} & \textbf{0.936} & \textbf{24.334} &\textbf{ 0.921} & \textbf{24.803} & \textbf{0.946} & \textbf{20.022} & \textbf{0.758}\\
\hline

\end{tabular}
}
\label{table_nighttime_dehaze}
\end{table*}

\paragraph{Compared Methods.}
For nighttime dehazing, we compare M\textsuperscript{3}KE with several specialized methods, including MRP \cite{Fast_Haze}, OSFD \cite{NightDehaze}, NightDeFog \cite{NightDeFog}, NightHazeFormer \cite{UNREAL}, NightEnhance \cite{NightEnhance}, SFSNiD \cite{SFSNID} and NightHaze \cite{NightHaze} for comparison. 
We also include daytime dehazing baselines for broader comparison, such as GridDehazeNet \cite{Griddehazenet}, FFA \cite{FFA_Net}, MSBDN \cite{MSBDN}, Dehamer \cite{Dehamer}, C2P \cite{c2p}, Fourmer \cite{Fourmer}, DehazeFormer \cite{DehazeFormer}, MITNet \cite{MITNet}, DEANet \cite{DEANet}, CORUN \cite{CORUN}, DCMPNet \cite{DCMPNet} and SGDN~\cite{sgdn}. 
For low-light enhancement, we evaluate against RetinexNet \cite{RetinexNet}, LightenDiffusion \cite{LightenDiffusion} and ReDDiT \cite{ReDDiT}.

\subsection{Qualitative Evaluation}
As shown in Figure~\ref{figure_night_time_dehazing}, we first evaluate the performance of our method on synthetic nighttime dehazing datasets. It is evident that our approach significantly surpasses existing methods in producing clearer dehazed images, recovering finer structural and textural details, and generating visually more natural and realistic results.

We then further validate the effectiveness of our method on real-world nighttime haze images, as illustrated in Figure~\ref{figure_nighttime_on_real}. Unlike synthetic data, real nighttime scenes present more severe challenges due to complex lighting conditions and unstructured haze patterns. Prior methods tend to fail under these conditions, often generating hazy, dark, or overly saturated outputs that fail to simultaneously remove haze and enhance illumination.

In contrast, our method delivers substantially cleaner and more visually pleasing results, effectively removing haze while enhancing dark regions to restore scene visibility. This demonstrates not only the robustness of our model under challenging real-world conditions but also its strong generalization ability to unseen image distributions. These results underscore the practical applicability of our approach in real nighttime scenarios.

\subsection{Quantitative Evaluation}
Table~\ref{table_nighttime_dehaze} summarizes the quantitative comparison results on nighttime dehazing datasets, including NHR \cite{NightDehaze}, NHM \cite{NightDehaze}, NHCL \cite{NightDehaze}, NHCM \cite{NightDehaze}, NHCD \cite{NightDehaze}, and UNREAL-NH \cite{UNREAL}.
Our method achieves the highest SSIM across all datasets with notable gains.
Specifically, on the NHM \cite{NightDehaze} and NHCL \cite{NightDehaze} datasets, our model’s SSIM surpasses NightEnhance \cite{NightEnhance} and SFSNiD \cite{SFSNID} by 0.01, respectively. On the UNREAL-NH \cite{UNREAL} dataset, our model achieves an SSIM of 0.758, surpassing SFSNiD \cite{SFSNID} by 0.029. Regarding PSNR, our method demonstrates a 1.23 dB improvement over DEANet \cite{DEANet} on the UNREAL-NH \cite{UNREAL} dataset and a 0.636 dB enhancement over NightEnhance \cite{NightEnhance} on the NHR \cite{NightDehaze} dataset. This further demonstrates the superior performance of our method on the UNREAL-NH \cite{UNREAL} dataset, where both dehazing and low-light enhancement are simultaneously required. 

These consistent improvements across multiple metrics and datasets clearly demonstrate the effectiveness and robustness of our method for nighttime dehazing. In particular, the strong performance on the UNREAL-NH dataset—where both haze removal and low-light enhancement are required—highlights our method’s superior ability to handle compound degradations. 

\begin{table}[t]
\centering
\caption{Results of the ablation study. The best results are highlighted in bold. A checkmark (\checkmark) indicates the use of our proposed module; otherwise, a base module of similar complexity is used.} 
\label{ablation}
{\fontsize{5.5pt}{7pt}\selectfont
\resizebox{\linewidth}{!}{
\begin{tabular}{c|c|c|c|c}
\hline
Number of Experts& FAR  & Expert & PSNR↑ & SSIM↑ \\
\hline
1\qquad1\qquad1 & \checkmark & \checkmark & 18.123 & 0.717\\
2\qquad2\qquad2 & \checkmark & \checkmark & 19.992 & 0.741\\
\rowcolor{gray!15}    
3\qquad3\qquad3 & \checkmark & \checkmark & \textbf{20.022} & \textbf{0.758}\\
4\qquad4\qquad4 & \checkmark & \checkmark & 19.911 & 0.748\\
\hline
3\qquad3\qquad3 &  &  & 19.123 & 0.732 \\
3\qquad3\qquad3 &\checkmark   &  & 19.976 & 0.731 \\
3\qquad3\qquad3 &   & \checkmark & 19.315 & 0.737 \\
\rowcolor{gray!15}    
3\qquad3\qquad3 &\checkmark   & \checkmark & \textbf{20.022} & \textbf{0.758} \\

\hline
\end{tabular}}}
\label{ablation_study}
\end{table}

\subsection{Ablation Study}
We conduct ablation studies on UNREAL-NH \cite{UNREAL} datasets to investigate the effectiveness of different modules.
Please refer to the supplemental for more results.

\paragraph{Expert Number Effectiveness.}
We first conduct an ablation study to investigate the impact of the number of experts in our framework.
As shown in Table~\ref{ablation_study}, using only a single expert results in suboptimal performance, suggesting that it is insufficient to model the diverse characteristics of different degradation types. When the number of experts increases to two, the model exhibits a significant performance improvement, demonstrating the benefit of specialization.
Introducing a third expert leads to further gains and achieves the best overall performance. However, adding a fourth expert slightly degrades the results, likely due to under-training or over-fragmentation of the data, indicating that three experts strike a good balance between capacity and generalization.
Additional ablation studies on different combinations of expert levels are provided in the supplemental material, further validating the design choices in our expert-based framework.

\paragraph{Module Effectiveness.}
We further conduct ablation studies on our proposed modules by incrementally integrating them into the model.
When the Frequency-Aware Router (FAR) is not used, we employ a standard two-layer MLP as the router. In the absence of the Expert Block, it is replaced with a residual block of equivalent parameter size to ensure a fair comparison.
As shown in Table~\ref{ablation_study}, introducing the FAR module leads to a substantial performance gain, underscoring the limitations of conventional MLP-based routers in modeling cross-domain degradation patterns. Furthermore, incorporating the Expert Block yields additional improvements by enabling mutual enhancement between the frequency and visual domains, demonstrating the effectiveness of our architectural design.

\begin{figure}[t]
\centering
\includegraphics[width=1\columnwidth]{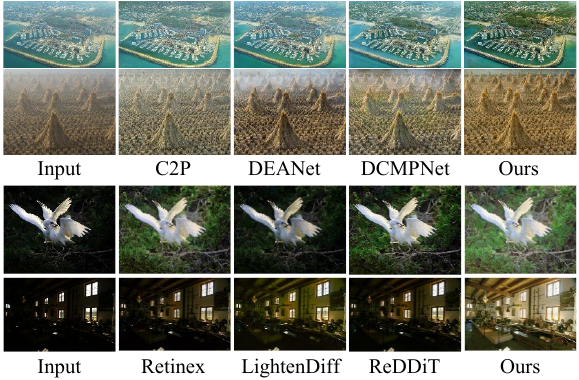} 
\caption{
Qualitative comparison of the proposed method with prior works for real-world daytime dehazing (first two rows) and low light enhancement (last two rows).}
\label{figure_daytime_lowlight_on_real}
\end{figure}

\begin{figure}[t]
\centering
\includegraphics[width=0.99\columnwidth]{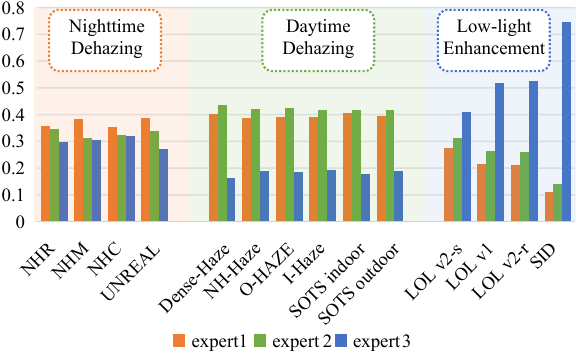
}
\caption{The visualization of the average weights assigned to the three image-level experts for each dataset.}
\label{figure_visualization_of_expert}
\end{figure}

\subsection{Discussions}
To further validate the superior performance and generalizability of our model, we first show the visual comparisons across diverse and hazy low-light scenarios, and then visualize the average weights assigned to the three image-level experts on a variety of datasets.
Please refer to the supplementary material for additional comparisons and discussions.

\paragraph{Daytime Image Dehazing.}
We further conduct a qualitative comparison on real-world daytime dehazing. Specifically, we compare our method with C2P~\cite{c2p}, DEANet~\cite{DEANet}, and DCMPNet~\cite{DCMPNet} on the Fattal \cite{fattal} dataset. As shown in the first two rows of Figure~\ref{figure_daytime_lowlight_on_real}, our method achieves superior visual quality, providing cleaner haze removal and more faithful restoration of scene details compared to prior approaches. 

\subsubsection{Low Light Image Enhancement.}
We further conduct a qualitative comparison on real-world low-light enhancement images. Specifically, we compare our method with RetinexFormer~\cite{Retinexformer}, LightenDiffusion~\cite{LightenDiffusion}, and ReDDiT~\cite{ReDDiT}. As shown in the last two rows of Figure~\ref{figure_daytime_lowlight_on_real}, our method delivers superior visual quality, achieving more thorough illumination enhancement compared to previous approaches. These results clearly demonstrate the effectiveness of our method on real-world data and highlight its strong generalization capability across diverse enhancement tasks. 

Overall, our model is capable of simultaneously handling image dehazing and low-light enhancement under a variety of challenging conditions, consistently achieving highly satisfactory results. 
From dense haze to extreme low-light conditions, our framework excels in restoring clarity and brightness while preserving natural color balance and fine details.

\subsubsection{Expert Assignment Weight Visualization}
Figure~\ref{figure_visualization_of_expert} visualizes image-level expert weights across datasets.
For each dataset, the average expert assignment weight is calculated and visualized.
We can see that images with different types and degrees of degradation are assigned to different experts. 
Overall, Experts 1 and 2 are better suited for daytime dehazing, and Expert 3 is more effective for low-light enhancement.
Nighttime dehazing shows balanced weights across all three experts, reflecting its hybrid degradation nature.
For example, in the densest haze dataset, DenseHaze \cite{DENSE-HAZE}, Experts 1 and 2 have the highest weight, which gradually decreases as haze density diminishes. Similarly, in low-light enhancement degradation, as the level of darkness increases, Expert 3’s weight rises, peaking in the SID \cite{SID} dataset, which has the most extreme low-light conditions. These results demonstrate that our router effectively distinguishes different input types, enabling each expert to specialize in its respective restoration task. Visualizations of expert assignments at the patch-level and pixel-level can be found in the supplementary material.

\begin{figure}[t]
\centering
\includegraphics[width=0.99\columnwidth]{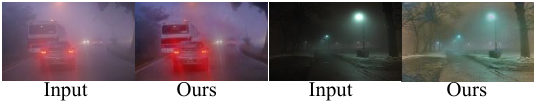}
\caption{Failure cases. Our model may fail on some images with extremely heavy glow or dynamic lighting.}
\label{failure}
\end{figure}

{
    \small
    \bibliographystyle{ieeenat_fullname}
    \bibliography{main}
}


\clearpage

\appendix

In this supplemental material, we provide further details and results to supplement the main paper. In Section~A, we elaborate on the architectural details of our framework. Section~B presents more comprehensive ablation studies on expert types and loss functions. In Section~C, we showcase additional results on real-world daytime hazy and low-light scenes, as well as on various test datasets. Finally, Section~D illustrates the expert alignment at both the patch and pixel levels.

\section{More Model Details}
\begin{figure*}[h]
\centering
\includegraphics[width=1.99\columnwidth]{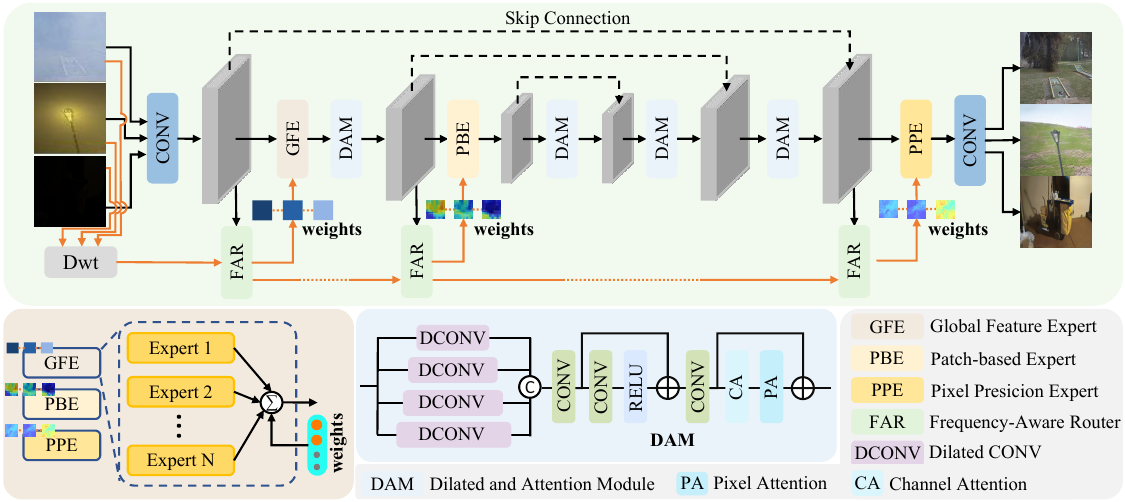}
\caption{Model details.}
\label{model details}
\end{figure*}

As illustrated in Figure~\ref{model details}, our M\textsuperscript{3}KE network utilizes a three-level encoder-decoder U-Net architecture. During each downsampling or upsampling step, the number of channels is doubled/halved, while the size of the feature map is halved/doubled. Skip connections link feature maps of equal sizes between the encoder and decoder to preserve the original image details. The encoder uses a 3×3 convolution with a stride of 2 for three downsampling operations. After the first downsampling, an image-based expert called the Global Feature Expert (GFE) is employed. After the second downsampling, a patch-based expert, referred to as the Patch-Based Expert (PBE), is used.

In the decoder, 4×4 transposed convolutions are utilized for upsampling. After the final upsampling, a pixel-based expert named the Pixel Precision Expert (PPE) is applied. As described in the main text, each expert is implemented using an Integrated Frequency Interaction Block (IFIB), and weights are assigned to different levels of experts using the Frequency-Aware Router (FAR).

In both the encoder and decoder, at each downsampling or upsampling stage, we incorporate a module consisting of multi-scale dilated convolutions along with channel and pixel attention, termed the Dilated and Attention Module (DAM). This module uses dilated convolutions with rates of 1, 2, 3, and 4 to extract receptive field information at different scales. The outputs are concatenated and passed through a conv-relu residual structure, followed by channel and pixel attention mechanisms to enhance feature representation, all while employing residual connections for stability.

We also plan to design dynamic network architectures \cite{ea-vit} that can adaptively adjust to low-level image restoration tasks, enabling more efficient and flexible image restoration methods. This direction will be explored in future work.

\section{More Ablation Studies}
In this section, 
we provide additional ablation experiments.
All the following ablation experiments are conducted using the same datasets as we use in the main text.
Notably, the SID \cite{SID} dataset used is a customized subset captured by the Sony $\alpha$7S II camera. The RAW images were converted to sRGB without gamma correction, resulting in exceptionally dark images.

\subsection{Expert Types and Combinations}
In our work, we use a sequence of experts starting with image-based, followed by patch-based, and finally pixel-based experts. This progression allows for initial coarse image restoration, followed by specific patch-level recovery that emphasizes degradation-specific mutual learning, and ends with fine-grained pixel-level refinement. However, to determine if this combination is optimal, we conduct the following experiments. We uniformly use three experts for each type and compare our approach with experiments using only image-, patch-, or pixel-based experts, as well as a different sequence starting with pixel, then patch, and finally image.

\begin{table}[h]
\fontsize{7pt}{9.2pt}\selectfont
\centering
\caption{Ablation study on expert types and combinations.} 
\tabcolsep=0.052\linewidth
\resizebox{\linewidth}{!}{
\begin{tabular}{ccc|c|c}
\hline
\multicolumn{3}{c|}{Experts Type}  & Avg PSNR↑ & Avg SSIM↑ \\
\hline
image & image & image&   22.861 & 0.862\\
patch & patch & patch&   22.153 & 0.856\\
pixel & pixel & pixel&   22.562 & 0.869\\
pixel & patch & image&   22.421 & 0.864\\
\hline
image & patch & pixel & 23.143 & 0.874\\
\hline
\end{tabular}}
\vspace{-0.1in}
\label{expert type}
\end{table}

As shown in Table \ref{expert type}, our experimental results demonstrate that our expert strategy achieves the best performance in both PSNR and SSIM metrics. Using only image-based experts results in relatively good PSNR performance but falls short in SSIM. The approach using only patch-based experts performed the worst, possibly because relying solely on patch-based processing is not optimal. Similarly, using only pixel-based experts or other combinations did not achieve results comparable to our proposed strategy.

\subsection{Loss Functions}
During model training, we used a combination of four different loss functions. To understand the contribution of each loss, we conducted the following experiments: progressively adding each loss function and observing the impact on the model's performance.

As shown in Table \ref{ablation loss function}, all four of our loss functions contribute effectively to the model's performance. L1 loss, commonly used in low-level vision tasks, performs well for general image restoration. SSIM loss significantly improves the structural similarity of the restored images, while perceptual and adversarial losses further enhance the overall performance of the model.

\begin{table}[h]
\fontsize{7pt}{9.2pt}\selectfont
\centering
\caption{Ablation study on loss functions.} 
\label{ablation}
\begin{tabular}{c c c c|c c}
\hline
 Smoth L1  & MS-SSIM & Perceptual & Adversarial &PSNR$\uparrow$ & SSIM$\uparrow$ \\
\hline
\checkmark &  &  &  & 22.452 & 0.851  \\
\checkmark   & \checkmark &  & & 22.841& 0.865 \\
\checkmark   & \checkmark &\checkmark  & & 22.940 & 0.869  \\
\checkmark   & \checkmark &\checkmark  &\checkmark &23.143 & 0.874 \\

\hline
\end{tabular}
\vspace{-0.1in}
\label{ablation loss function}
\end{table}

\begin{table*}[h]
\small
\centering
\caption{Quantitative comparison on unpaired low-light image enhancement datasets.
The best result is marked in bold, and the second-best is marked with an underline.
}
\fontsize{7pt}{9.2pt}\selectfont
\resizebox{\linewidth}{!}{
\begin{tabular}{c c|c c|c c|c c|c c|c c}
\hline
\multicolumn{2}{c|}{\multirow{2}{*}{Method}} &\multicolumn{2}{c|}{DCIM }&\multicolumn{2}{c|}{LIME}& \multicolumn{2}{c|}{MEF }&\multicolumn{2}{c}{NPE }&\multicolumn{2}{c}{VV }\\
\cline{3-12}
\multicolumn{2}{c|}{} & BRISQUE$\downarrow$ & NIQE$\downarrow$ & BRISQUE$\downarrow$ & NIQE$\downarrow$ & BRISQUE$\downarrow$ & NIQE$\downarrow$ & BRISQUE$\downarrow$ & NIQE$\downarrow$& BRISQUE$\downarrow$ & NIQE$\downarrow$\\
\hline

\multicolumn{2}{c|}{RUAS \cite{RUAS}} &38.75 &5.21 &27.59 &\underline{4.26} &\underline{23.68} & \underline{3.83} &47.85 & 5.53&38.37&4.29 \\ 
\multicolumn{2}{c|}{KinD \cite{KinD}} &48.72 &5.15 & 39.91&5.03 &49.94 &5.47 &36.85&4.98&50.56&4.30 \\ 
\multicolumn{2}{c|}{LLFlow \cite{LLFlow}} &\textbf{26.36} &4.06 &27.06 &4.59 &30.27 &4.70 &28.86 &4.67&\textbf{31.67}&4.04 \\
\multicolumn{2}{c|}{SNR-Aware \cite{SNRAware}} &37.35 &4.71 &39.22 &5.74 &31.28 &4.18 &\underline{26.65} &4.32&78.72&9.87 \\
\multicolumn{2}{c|}{PairLIE \cite{fu2023learning}} &33.31 &\underline{4.03} &\underline{25.23} &4.58 &27.53 &4.06 &28.27 &\underline{4.18}&39.13&\underline{3.57} \\

\hline
\multicolumn{2}{c|}{Ours} &\underline{29.09} &\textbf{3.57} &\textbf{20.54} &\textbf{3.61} &\textbf{17.56} &\textbf{3.67} &\textbf{25.31} &\textbf{3.68} &\underline{32.56} & \textbf{3.40} \\ 
\hline
\end{tabular}}
\label{DCIM,LIME,MEF,NPE and VV}
\end{table*}

\section{More Results}
To further evaluate the performance of our model in real-world scenarios, we selected several real-world hazy and low-light datasets to compare against existing methods.

\begin{figure*}[h]
\centering
\includegraphics[width=1\textwidth,height=1\textheight]{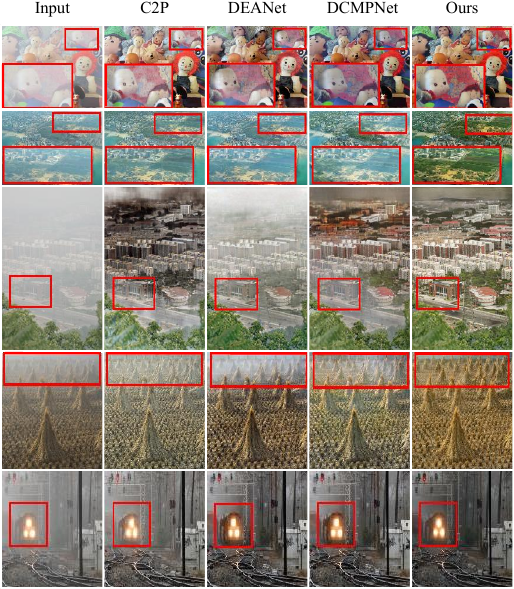}
\caption{
Qualitative comparison of real-world daytime dehazing.}
\label{figure_day_time_realworld_dehazing}
\end{figure*}

\subsection{Visual Results on Real Hazy Scenarios}
For Dehazing, we utilized the Fattal dataset \cite{fattal} and randomly selected some dehazed images to compare with other approaches. As illustrated in the Figure \ref{figure_day_time_realworld_dehazing}, our model demonstrates superior performance on real-world hazy images. Compared to other dehazing methods, our approach achieves more thorough haze removal, resulting in images that appear more natural without color distortion.

\subsection{Visual Results on Real Low-Light Scenarios}
\begin{figure*}[h]
\centering
\includegraphics[width=1\textwidth,height=1\textheight]{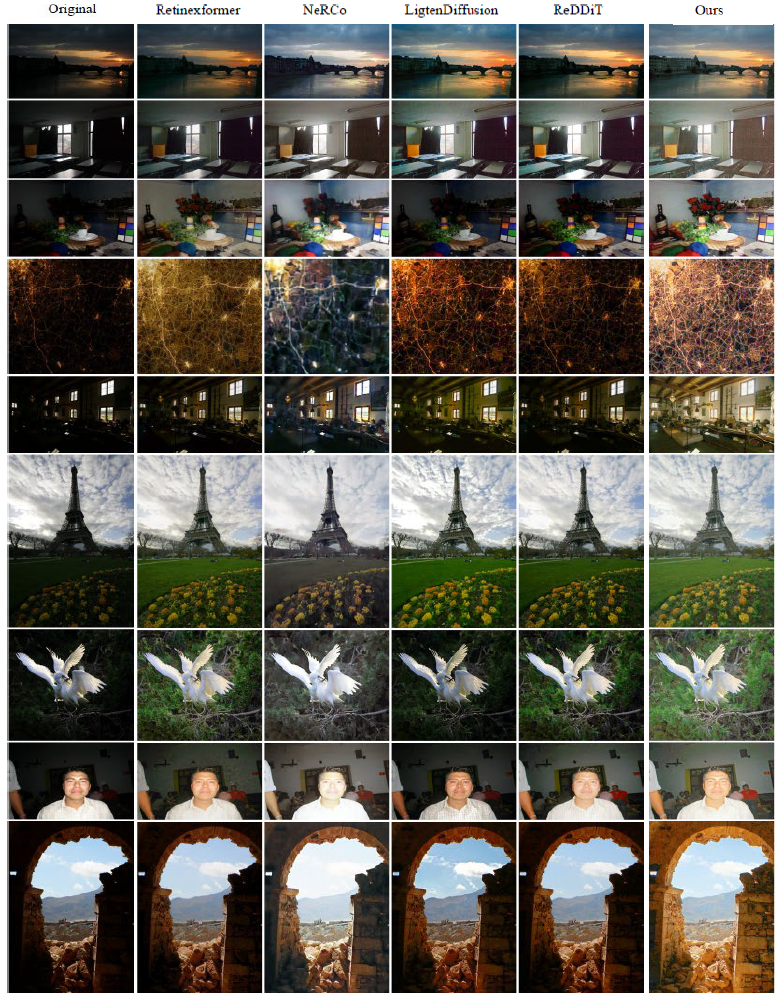}
\caption{Qualitative comparison of our method with other approaches on the real-world low light enhancement dataset Fattal.}
\label{realworldlow}
\end{figure*}

For Low-Light Enhancement, we selected five commonly used benchmarks: DICM \cite{DICM}, LIME \cite{LIME}, MEF \cite{MEF}, NPE \cite{NPE}, and VV \cite{VV}. Similarly, we randomly selected some images for comparison against other methods. As shown in the Figure \ref{realworldlow}, our approach delivers more effective brightness enhancement, producing images with more natural brightness levels and well-recovered details, which significantly surpasses previous methods.

\subsection{More Results on Test Datasets}
\begin{figure*}[h]
\centering
\includegraphics[width=1\textwidth,height=1\textheight]{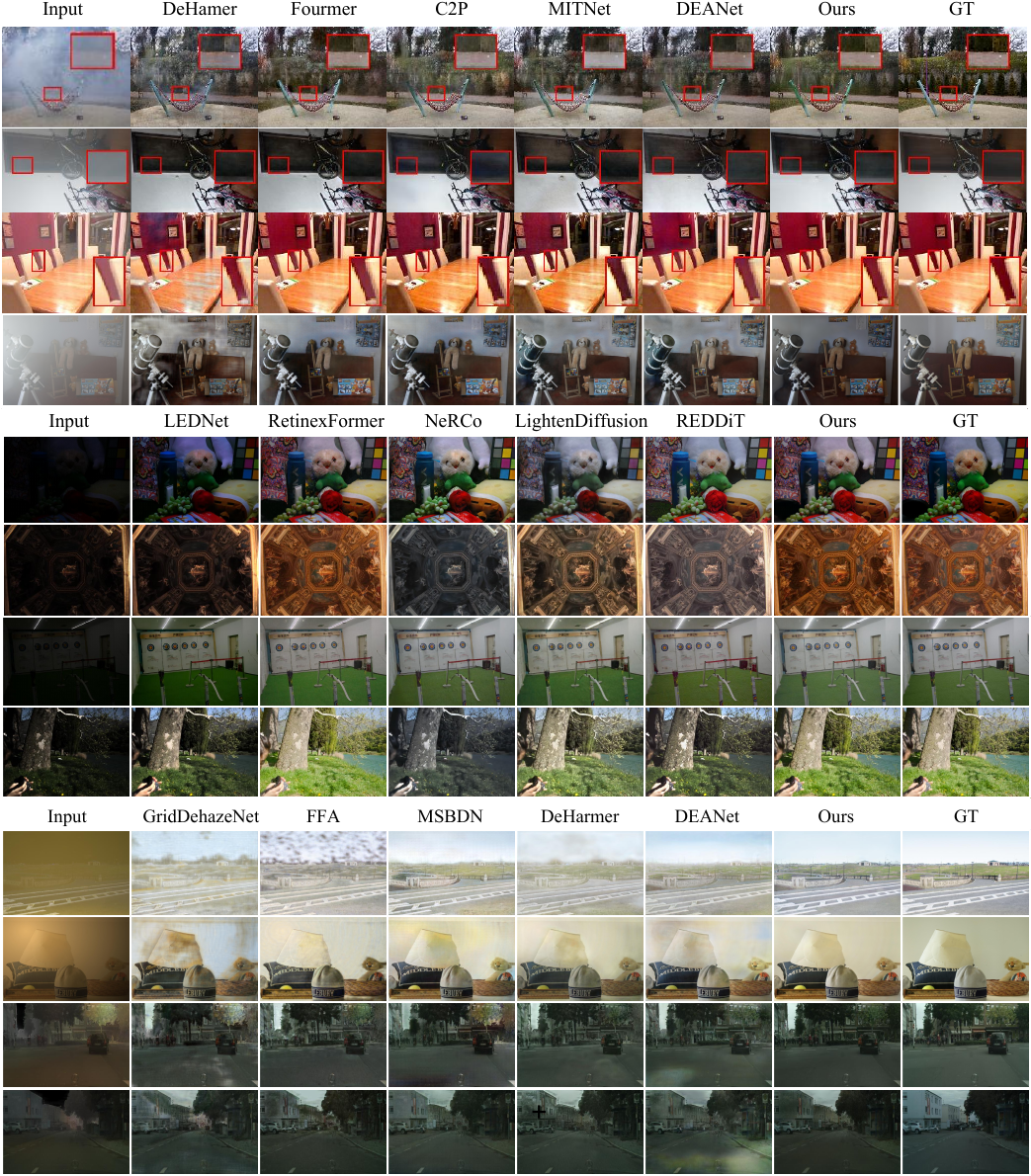}
\caption{Qualitative comparison of our method with other approaches on datasets atasets for synthetic nighttime dehazing, daytime dehazing, and low-light
enhancement.}
\label{someotherdatasets}
\end{figure*}

As shown in the figure \ref{someotherdatasets}, we present additional results on datasets for synthetic nighttime dehazing, daytime dehazing, and low-light enhancement.

These results validate our model's adaptability to various image distributions, showcasing its potential for practical real-world applications and significantly enhancing its applicability across diverse scenarios.
t



\section{Expert Assignment Weight Visualization}
In the main paper, we presented the expert assignment at the image level. Here, we demonstrate the expert assignments at the patch and pixel levels.

\subsection{Patch-Level Expert Assignment}
\begin{figure*}[h]
\centering
\includegraphics[width=1\textwidth,height=1\textheight]{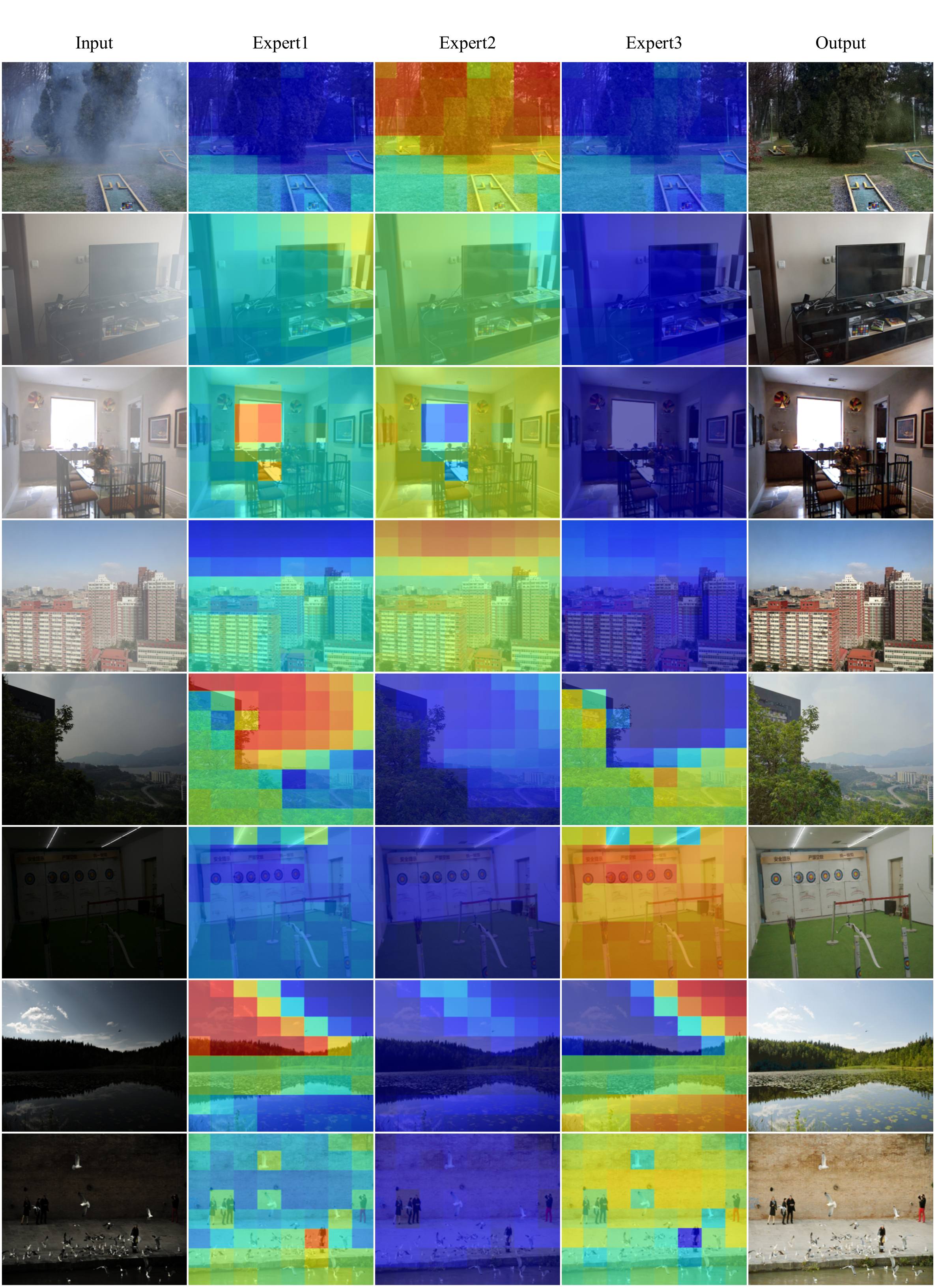}
\caption{Visualization results of three experts on Patch-level.}
\label{patch_level}
\end{figure*}

As illustrated in the figure \ref{patch_level}, the patch-based expert assignment is highly correlated with the degraded regions in the current image. For example, in the case of dehazing, non-uniform hazy images exhibit distinct expert weights for hazy and non-hazy patches. Similarly, for outdoor hazy images, the expert weights assigned to sky regions differ significantly from those assigned to non-sky regions. Additionally, the background color of objects also affects patch weights.
For low-light enhancement, a similar pattern is observed for non-uniform lighting conditions, such as in areas with light sources. The expert weights in these regions differ noticeably from those in standard low-light areas.

It can also be observed that different experts are primarily used for dehazing and low-light tasks. Dehazing datasets mostly utilize Expert 2, while low-light datasets primarily use Experts 1 and 3. However, there are also cases where both tasks share experts extensively, demonstrating the effectiveness of the mutual use of experts.

\subsection{Pixel-Level Expert Assignment}
\begin{figure*}[h]
\centering
\includegraphics[width=1\textwidth,height=1\textheight]{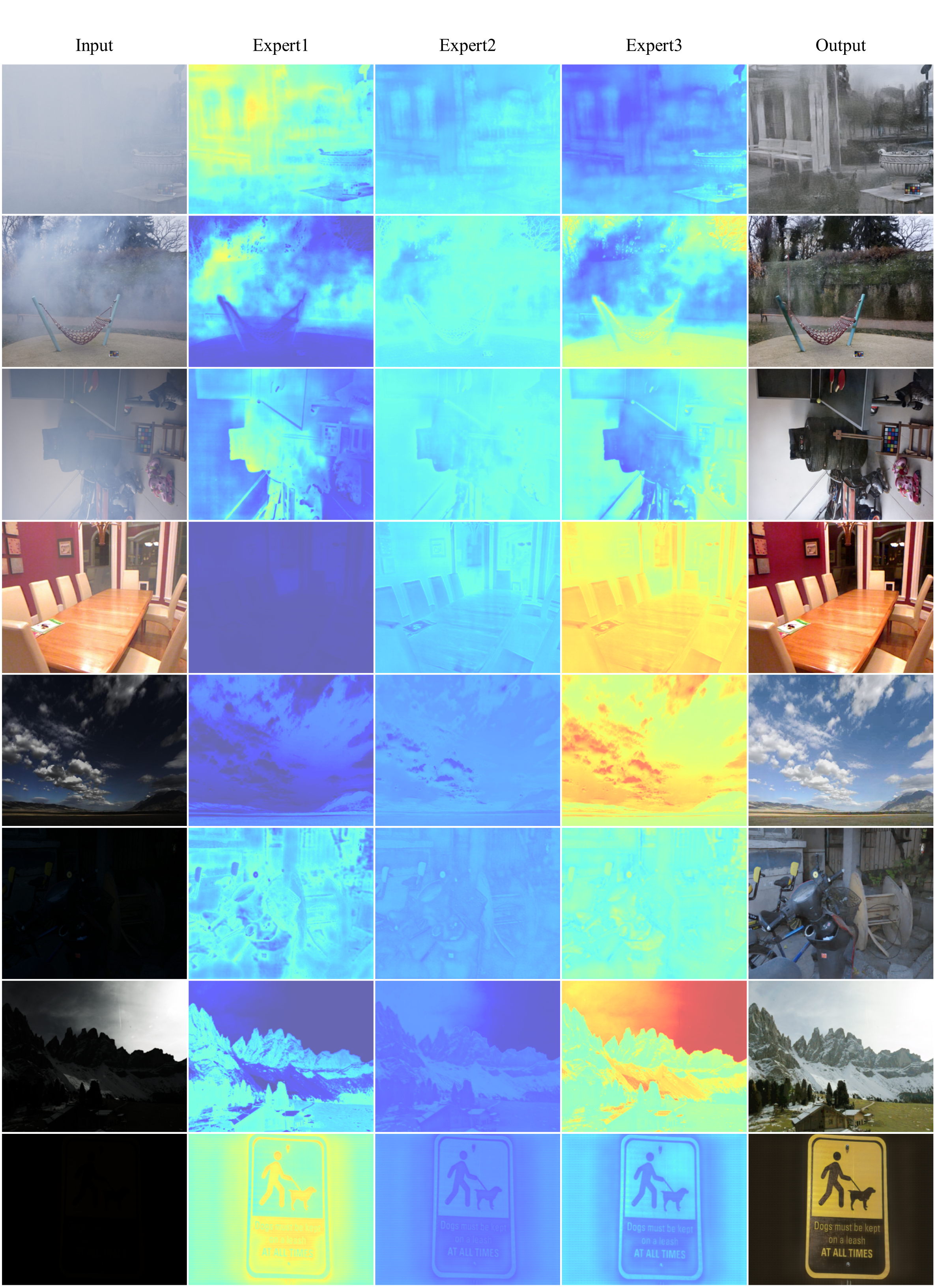}
\caption{Visualization results of three experts on Pixel-level.}
\label{pixel_level}
\end{figure*}

As shown in the Figure \ref{pixel_level}, unlike the patch-based expert assignment, where one or two experts dominate the weights for each image, the pixel-based expert assignment tends to distribute weights more evenly across all three experts. From the heatmap, it is clear that pixel-level expert assignment effectively highlights both the low-frequency regions and high-frequency edges of the image. This demonstrates that the pixel-level experts excel at handling non-uniform conditions and restoring high-frequency details.

\end{document}